\documentclass[letterpaper, 10 pt, conference]{ieeeconf}  

\IEEEoverridecommandlockouts                              

\title{\LARGE \bf
Crowd against the machine: A simulation-based benchmark tool to evaluate and compare robot capabilities to navigate a human crowd
}

\author{Fabien Grzeskowiak$^{1}$, David Gonon$^{2}$, Daniel Dugas$^{3}$, Diego {Paez-Granados}$^{2}$, Jen Jen Chung$^{3}$, Juan Nieto$^3$\\ Roland Siegwart$^{3}$, Aude Billard$^{2}$, Marie Babel$^{1}$ and Julien Pettr{\'e}$^{1}$ 
\thanks{$^{1}$F. Grzeskowiak, M. Babel and J. Pettr{\'e} are with Univ Rennes Inria CNRS IRISA, Rennes, France. {\tt\small julien.pettre@inria.fr}}%
\thanks{$^{2}$D. Gonon, D. {Paez-Granados} and A. Billard are with EPFL, Lausanne, Switzerland. {\tt\small aude.billard@epfl.ch}}%
\thanks{$^{3}$D. Dugas, J. J. Chung, J. Nieto and R. Siegwart are with ETH, Zürich, Switzerland. {\tt\small jenjen.chung@mavt.ethz.ch}}%
}
\usepackage{url}
\usepackage{graphicx}
\usepackage{color}
\usepackage[space]{grffile}
\usepackage{multirow}

\newcommand{\anon}[1]{#1}

\begin{document}

\maketitle
\thispagestyle{empty}
\pagestyle{empty}

\begin{abstract}

The evaluation of robot capabilities to navigate human crowds is essential to conceive new robots intended to operate in public spaces. This paper initiates the development of a benchmark tool to evaluate such capabilities; our long term vision is to provide the community with a simulation tool that generates virtual crowded environment to test robots, to establish standard scenarios and metrics to evaluate navigation techniques in terms of safety and efficiency, and thus, to install new methods to benchmarking robots' crowd navigation capabilities. This paper presents the architecture of the simulation tools, introduces first scenarios and evaluation metrics, as well as early results to demonstrate that our solution is relevant to be used as a benchmark tool. 

\end{abstract}

\section{INTRODUCTION}

A large number of mobile robots are conceived with the intention of being operated in public spaces, likely to be frequented by crowds. Assessing robots' ability to navigate through the crowd in a way that is efficient and safe is thus of paramount importance. However, such assessment is difficult for several reasons. Firstly, a crowd is a dynamic and complex environment, the state of which depends on a very large number of parameters. A meaningful evaluation should cover as much of this parameter space as possible, this guarantees that the robot behaviour has been studied in most of the situations it may encounter. However, this crowd complexity compounds the already high-dimensional parameter space related to the robot behaviour. 

Ideally, we should study and test various robot behaviours in the same situations. This would enable comparing efficiency and safety levels offered by various navigation techniques. Unfortunately, this results in a combinatorial explosion. The risks linked to the participants and the cost of conducting real-world experiments are our two last major problems. This explains why research work dedicated to robot navigation in crowds are limited to tests with a limited number of participants that barely constitute a crowd. To mitigate this issue, our paper explores solutions to provide the community with a crowd-robot navigation benchmark tool. Our work addresses three aspect of this problem: i) to define reference navigation scenarios, ii) to provide simulation algorithms and a framework for the simulation itself, iii) to introduce metrics for evaluating efficiency and safety levels. 

Simulation-based evaluation offers valuable insights. Nevertheless, it cannot replace the validity of real tests because, if robot simulation cannot already be considered perfectly accurate, the simulation of human behaviour is even more complex. It is impossible to simulate all the behaviour that a crowd can exhibit, in particular, its reactions to a robot. However, crowd simulation is useful to create a synthetic test environment. It can be a preliminary and complementary tool to real tests that evaluate the evolution of a robot in an environment that is representative in many aspects to a real crowd. 

At the very least, simulation-based testing addresses the four problems outlined above: it can automatically explore large situational parameter spaces, it eases comparisons and is, without question, safe. Moreover, simulation brings the major interest of allowing one to easily define a set of \emph{standard situations}, which are defined by the configuration of the crowd and the task of the robot. These can be precisely replicated from one series of tests to another, which can be elaborated and shared at the scale of a whole community. It therefore makes simulation-based testing an ideal candidate to be used as a benchmark tool to fairly evaluate and compare the behaviour of robots in a crowd.



Our article presents the following contributions: (i) a simulation platform, open and accessible to the community, allowing the simulation of robot motion among a moving crowd, (ii) a preliminary set of standard situations, representative of common ones to which a robot is exposed when navigating in a crowded environment, (iii) a set of metrics to evaluate the test results with regard to the robot's efficiency, the disturbance caused to the crowd's navigation, as well as the related risks of collisions, (iv) a demonstration of this tool's capabilities to be used as a benchmark. 

\section{RELATED WORK}

The development of new mobile robots intended to move within crowds requires the evaluation of their navigation capabilities in terms of efficiency and safety. Simulators facilitate such evaluation by providing interfaces through which a robotic component can manipulate and perceive an abstract environment that models a real phenomenon (in our case, a crowded environment). 
In the following, we review some existing crowd simulators by investigating which technical concepts and metrics they implement that make them significant for robotic navigation in human crowds.

A contribution of \cite{8403274} is a probabilistic planner that updates its belief on pedestrians' goal positions and assumes that their decision making is based on PORCA, a variant of ORCA \cite{van2011reciprocal} (using a different cost function modeling pedestrians' impatience). Thus, the simulator is suitable to demonstrate exactly this planner's performance as the planner's underlying assumptions about pedestrians' decision making match exactly how the simulation controls them. The metrics for the evaluation are the robot's collision rate, travel time, length of acceleration/deceleration periods, and success rate (to reach the goal). 
\cite{6942731} compare how a robot using different learning-based controllers performs when navigating among pedestrians in the PedSim\footnote{\url{http://pedsim.silmaril.org/}} simulator. They measure the robot's path length and smoothness as well as pedestrian comfort by the number of proxemic intrusions by the robot (proximity at particular angles).
\cite{8794134} evaluate their navigation technique for the robot in simulations with agents that apply ORCA \cite{van2011reciprocal}. They report the robot's success rate (to reach the goal without a collision), collision rate, time, and reward (referring to their framework of reinforcement learning).
\cite{kuderer2012feature} propose an approach that uses inverse reinforcement learning and predicts collective trajectories by assuming they optimize an underlying reward function. To show that their approach works as expected, they perform simulations wherein such a reward function governs the agents' behaviour, and the approach then learns this function to make reliable predictions.
\cite{doi:10.1177/0278364914557874,trautman2010unfreezing} call their experiments simulations wherein they replay crowd motions from a recorded dataset and replace just one pedestrian by the robot who then needs to plan its own path (whereas others do not react to it). They measure the robot's minimum distance to pedestrians and the robot's path length.
Similarly, \cite{8263556} use the same dataset and crowd replay to simulate how the robot reacts to pedestrians (while they do not react to it).
Webots\footnote{\url{https://www.cyberbotics.com/}} is an open-source simulator offering animated human characters, albeit looking and walking rather like puppets. \cite{fraichard2020crowd} perform crowd simulations which incorporate heterogeneous navigation methods and limited perception to investigate the difficulties that one needs to tackle to enable robots to navigate in crowds. In Webots, they simulate how a crowd with limited field of view and different control laws performs, measuring the agents' number of collisions and time to reach their goals.

Some work aiming towards robotic navigation in crowds considers as the key task to understand and accurately model pedestrian decision making. One could argue that a robot just needs to replicate the human decision process to navigate among pedestrians like a regular pedestrian. They typically do not use simulation to evaluate their methods. Rather, they (e.g. \cite{turnwald2016understanding}) measure the similarity between the trajectories which their methods generate for the robot in a given situation to the motion of a real human subject in the same situation (also like in \cite{doi:10.1177/0278364914557874}, \cite{trautman2010unfreezing}).
The evaluation in \cite{turnwald2019human} lets human participants interact with a virtual pedestrian and judge later on whether the virtual pedestrian was following another human participant's decisions or an automatic game theoretic planner. This varied Turing test could show whether or not the decisions made in the background by the planner or the second participant are indistinguishable.

In summary, existing works on robots navigating in crowds mostly just consider one respective facet of crowds in their simulations, which is likely representing the specific task they are addressing. Therefore, it seems that a more versatile simulator, which reflects and integrates such aspects, could give new impulses to similar research. In particular, the above simulators are limited to local collision avoidance, i.e. do not exhibit high-level path planning, and they mostly lack realistic gait animation and graphical rendering of pedestrians, while realistic visual rendering is important for fully vision-based navigation techniques (e.g. end-to-end learning). Further, their metrics are mostly the same, surprisingly not quantifying the crowd's efficiency in most evaluations. The metrics for safety also seem fairly simplistic with most only considering the robot's distance to pedestrians. Colliding agents' mass, velocity and kinetic energy are suitable to quantify the potential for injury as several works in human-robot interaction have shown~\cite{7353110},~\cite{doi:10.1177/0278364912462256}.

\section{CROWDBOT SIMULATION PLATFORM}

\begin{figure}
  \includegraphics[width=\linewidth]{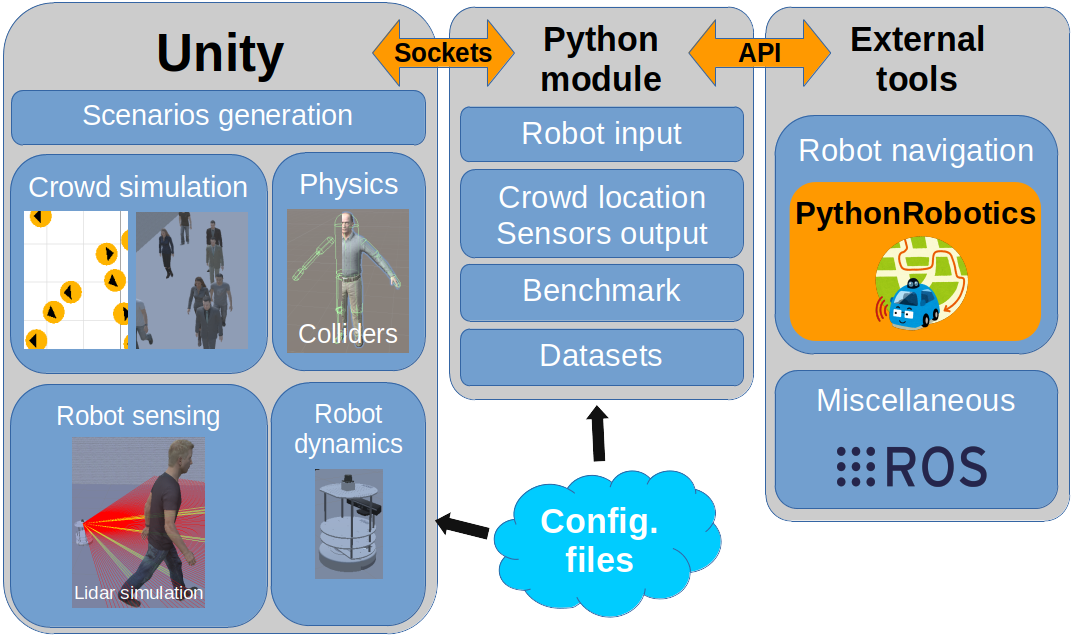}
  \caption{\label{fig:architecture} \textbf{The \anon{CrowdBot} simulator architecture:} Part of the tool is an application based on Unity (\protect\url{https://unity.com}). This application can generate scenarios, it simulates how the crowd moves and renders it in a 3D environment. It also handle physics simulation in order to compute collisions between the crowd and the robot. Finally, it handles the robot dynamics and sensing. A second part of the tool is the python module which controls the Unity application. This module conducts the simulation with instructions and data sent to the application through sockets, which then sends back simulation data. This module helps in the generation of datasets of crowd and robot trajectories, which can be analyzed using the benchmark tools. Finally, the module propose an API for ROS and PythonRobotics. One last component is the set of configuration files which gives the parameters of the simulation.}
\end{figure}


This section describes the \anon{CrowdBot} simulation architecture shown in Fig.~\ref{fig:architecture}.



\subsection{Crowd simulation}
The purpose of the crowd simulation component is to compute the motion of a crowd in a given environment, represented as a collection of moving points, one for each simulated individual. For this, we rely on microscopic crowd simulation algorithms, which perform this calculation based on the notion of an agent whose trajectory is a function of its goals and its interactions with the environment. Note that literature has proposed a number of numerical models of local interactions, and more specifically of collision avoidance, that determine agents' behaviours \cite{helbing2002simulation} \cite{van2011reciprocal} \cite{dutra2017gradient}.

To avoid the effect of limiting crowd behaviours to one of a unique algorithm, we employ a recent crowd simulation technique which is capable of reproducing a collection of crowd simulation algorithms \cite{van2020generalized}. The motion of each agent is driven by the definition of a cost function that the agent optimizes, defined in the agent's velocity space, which depend on the state of the agent as well as the state of one of its neighboring objects (other agents, robot, obstacles). Several existing algorithms can be reproduced by changing the definition of this cost function. In addition, agent motion is subject to the parameter settings of this function.

By choosing such a technique to simulate crowds in the CrowdBot framework, we open the possibility to include, in experimental plans, a variety of simulation algorithms as well as settings.



\subsection{Environment and human simulation}
Robot simulations are usually based on 3D rendering engines with physics simulation, in order to reproduce the robot's real environment.
In this paper, we focus on mobile robots that operate on the $x$-$y$ plane, defining the robot workspace on which dynamic obstacles can interact with the robot.
For our crowd simulation dedicated to robotics applications, we extend the 2D representation of the crowd with animated avatars walking according to the crowd simulation output. Those avatars are defined with a visual and a collision model, used by the physics engine, to interact with the robot.
We use Unity3D in order to have high definition rendering with physics simulation.
The crowd simulation library UMANS \cite{van2020generalized} is integrated in Unity thanks to a dedicated interface (API).
The simulation, running in Unity, can be controlled externally thanks a ROS node, or through a dedicated python module.

\subsection{Robot simulation}
In our simulator, a mobile robot is defined as a mechanical model associated to a visual model, as well as a linear and angular velocity command as input to control the robot motion, and outputs that sense the robot' environment.
The visual model is a list of 3D models of the joints of the robot associated with a skeleton of the robot, which helps defining the mechanical behaviour of the robot. 

Our simulator offers various ways to define the mechanical properties of a robot.
First, every robot we implemented can be considered as a kinematic rigid body. This method is useful for simple simulations when physics is not required, and can be used on every robot. The second method consist in considering the robot as a unique point-mass, or a set of connected rigid bodies. With this method, the robot is moved around using forces applies on the base of the robot. Friction limits the mobility of the robot, uniformly or non-uniformly, which is useful to define holonomic robots as well as differential drive robots or car-like robots.
The third way to control a robot, dedicated to wheeled robots, is through giving velocity commands to a regulator which directly controls the torque of the simulated motors in the wheels of the robot.

The robot model also defines a set of sensors, which uses the physics engine and the graphics rendering of Unity to generate outputs. The software can simulate various standard sensors: LiDAR, RGBD camera, ultrasound sensors, odometry. Sensor models also contain parameters that can be used to fit any sensor specification. Imperfections, such as noise and missing data, can be added to any sensor for more realism. The simulation also outputs information about the simulation, such as simulation time, physics engine report on agent-robot collision, the crowd position, or crowd mask for LiDAR, which gives the ID of the agent of the crowd detected in a LiDAR scan.

\subsection{Robot navigation plugin}
We made a simple python module in order to control the simulation.
The module sends data to the simulator: the simulation clock, which controls the time between simulation steps, the simulation controller, which is used to switch between scenarios or to stop the simulation, and a velocity command, directly used by the robot controller in the simulation.
When a step is executed by the simulation, the module receives all available data for this step: robot sensor outputs, crowd location and collision report.
We also provide a ROS node that uses this python module to control the simulation through ROS topics and ROS parameters and converts the received data into ROS topics.
The module is also designed to be used with PythonRobotics \cite{sakai2018pythonrobotics}, a python code collection for robotics algorithms, or more generally any python implementation of robotics algorithms.

\subsection{Initialisation pipeline and outputs}
The simulation require a extensive amount of parameters, stored in scenarios files.
Scenarios files are read by the Unity simulator and for a given simulation, a scenario file defines the 3D environment, the camera controls (third person view), the robots to load and their initial location, the crowd initial location, the density of the crowd, the flow of the crowd, i.e the set of goals for each agent, the path planner to use for each agent, the crowd reactivity to the robot (reactive, not reactive), and the crowd simulator (UMANS) basic parameters such as preferred speed or maximum acceleration. The scenario will use the default configuration of UMANS, but it can be be easily changed to any of the methods proposed by UMANS in another file.
Finally, some parameters are given as input for the python module directly or as ROS parameters if the ROS node is used: the time step of the simulation, the sleep time between steps (for real-time simulation), the ending conditions, and the number of scenarios to run. Also, the python module includes a recorder that saves JSON files of the whole simulation, which is configurable in the python module (file name and location or data to save).

\subsection{Open source project}
The benchmark tool is available freely for anyone on \url{ http://CrowdBot.eu/CrowdBot-challenge/}. On this website, one can download the CrowdBot challenge software which was used to generate the dataset and the benchmark results. It is possible to request access for the source code by contacting the developers on this page. This website also links to the documentation (wiki) of the software which gives details on the functionalities, tutorials and examples, and videos.

\section{CROWD ROBOT NAVIGATION BENCHMARK}

Our intention is to use the CrowdBot simulation platform as a benchmark tool to evaluate and compare robot crowd navigation capabilities in terms of safety and efficiency. This section presents both some navigation techniques we have selected from literature to be tested, as well as our evaluation protocol. 

\subsection{Navigation techniques and hypotheses} 

We have selected the 3 following navigation techniques: i) a ``Baseline" method, which consists in having the robot going straight toward the goal ignoring the crowd, ii) the dynamic window avoidance method (DWA) \cite{fox_dynamic_1997}, and iii) the reciprocal velocity obstacles method (RVO) \cite{van2011reciprocal}. The reason for this choice is that the robot will respectively: i) ignore the crowd agents, ii) consider them as static obstacles, and iii) predict the short term future motion of agents. With this choice, the benchmark tool shall demonstrate its capability to reveal benefits and drawbacks on the robot navigation efficiency and safety with respect to crowd agents. If our evaluation tool is adequate to be used as a benchmark, the evaluation should satisfy the following hypotheses:

\begin{itemize}
    \item[$H1$:] The Baseline method should get the best score in terms of efficiency, since the robot is ignoring the crowd. However, it should be the worst in terms of safety, since it shall move closer to crowd agents and provoke many collisions with them, not even trying to lower the collision forces with the crowd.
    \item[$H2$:] The DWA method, in comparison with Baseline, should show a lower number of collision, and thus a higher safety level. As the method considers obstacles to be static, it should show quite poor results in terms of efficiency, since it may put the robot on useless or late avoidance trajectories 
    \item[$H3$:] The RVO method should show the best compromise between efficiency and safety, as it is capable of predicting agents' motion and expects contribution from agents in performing avoidance.
\end{itemize}




\subsection{Benchmark setup}
\label{benchmark}


The idea of the benchmark is to propose a number of scenarios that each define the robot, the task of the robot and the activity of the crowd, with a number of parameters. The simulation platform generates trajectories for both the robot and the crowd, which are ultimately evaluated according to several metrics to compare the performances of various navigation techniques in similar conditions. In this paper, we illustrate the benchmark with a limited set of scenarios, which can be extended at will. 


\paragraph{Robot} We choose to illustrate the benchmark on one single robot, the TurtleBot2, a common differential drive robot. The simulated TurtleBot has a maximum velocity of 1 ms$^{-1}$ and a maximum acceleration of 5 ms$^{-2}$. It is controllable with a linear and angular velocity command. The simulation provides the crowd location, the robot odometry and the static obstacles. Those data can be used as an input for the robot navigation. The virtual robot is also equipped with a set of virtual sensors that offer the possibility of testing the robot navigation robustness to limited data: two 2D LiDARs that provide 360$^\circ$ coverage, a front-facing RGBD camera and 12 simulated ultrasound sensors.

\paragraph{Environment and task} the environment is defined as a corridor of 50m by 10m. The robot starts at one side of the corridor and its task is to reach the other side. The goal is reached when the robot travels 40m in the direction of the corridor. We enforces a time limit of 180 seconds to avoid infinite simulation for situations where the navigation does not find a solution. The robot is free to move in this corridor but cannot cross the borders delimited by four walls. The crowd however, can cross the walls. In the case that they traverse a wall, the agent's position is reset at the opposite one of the environment, which maintains a constant level of average density over the entire trial. 

\paragraph{Crowd activity} Our scenarios are finally defined by a crowd flow direction, a crowd behavior, and a crowd density. The possible crowd flows can be in the main direction of the corridor, pointing toward the same wall as the robot (1D+) or pointing toward the opposite wall, facing the robot (1D-). The crowd flow can also be bidirectional, in the main direction of the corridor, where half of the crowd is moving in the opposite direction of the other half (1Dx). Finally, the crowd can move in the secondary direction of the corridor, crossing the robot path, either in one main direction (2D), or with two opposing flows (2Dx). We selected two main crowd simulation algorithms which are fairly common and easily configurable: ``RVO", and ``Social Forces". These two algorithms can be configured as reactive or not reactive to the robot. We also chose to set the parameters of RVO with 0.5s of horizon time (the crowd only reacts when close to the robot) or with 1.5s horizon time, which results in more anticipatory interactions. Finally, we selected four levels of crowd density: 50 (0.1pm$^{-2}$), 100 (0.2pm$^{-2}$) , 200 (0.4pm$^{-2}$) or 350 (0.7pm$^{-2}$) agents in the corridor. With all those parameters combined, we end up with 100 scenarios. We randomly selected the initial position of the agents of the crowd and generated the definitive scenarios files to be used by the different robot navigation techniques.

\subsection{Evaluation metrics}
\label{metrics}
For this paper we limited the study to seven metrics which can be split into three categories: path efficiency, effect on the crowd flow, and crowd proximity. The latter two can be regrouped under the topic of crowd safety.
In the path efficiency metrics, we compare the situations where the robot is alone to the configuration where the robot is surrounded by a crowd. For those two configurations, we compute and compare the time taken by the robot to reach the goal ($T/Tcr$), the length of the robot path ($L/Lcr$), and the variation of speed (linear and angular) of the robot during the whole scenario ($J/Jcr$).
The metrics dealing with the effects on the crowd compare the speed of the robot neighbors (within 1m range) to the speed of the whole crowd ($NBR vel. = \frac{V_{Neighbors}}{V_{All}}$). The slower the neighbors are, the smaller the metric is.
In the second metric, we compare the angular velocity of the whole crowd to the velocity of the neighbors of the robot in scenarios ($NBR reac. = \frac{\omega_{All}}{\omega_{Neighbors}}$). If the neighbors rotates more than the rest of the crowd, $NBR reac.$ will be small.

Finally, the metrics dealing with crowd proximity are first the proximity metric ($Prox.$) which is defined by \[ Prox = 1 - \frac{1}{t_{final}}\sum_{t=0}^{t_{final}} d_{min}(t)/R \]
where $d_{min}$ is the distance to the closest agent, and R is the range in meter. In our case, we consider that there is at least 1 person that is in a 5 meter range near the robot.
The last metric is the colliding metric, which is defined by $Colliding = 1 - \frac{T_{Collision}}{T_{Scenario}}$ where $T_{Scenario}$ is the total time of the scenario, and $T_{Collision}$ is the cumulation of the collision instants. A score of 1 means no collisions at all.

\subsection{Collision Assessment}
We assess each collision by exploiting the simulator's report of the colliding human body part in order to compute its reflected mass $ m_{ref} $, which approximates the human inertia's contribution to the impact's kinetics. Each colliding body part's $ m_{ref} $ is computed as the reflected mass of the part's corresponding segment of a model that groups the human body into a vertical articulated chain of four rigid bodies, subsuming respectively the feet, lower legs, upper legs, and remaining upper body. These segments' lengths and inertial properties are sums of the subsumed parts' values for average adults (according to~\cite{doi:10.1002/9780470549148.ch4}). Following the approach in~\cite{doi:10.1177/027836499501400103}, we compute for our model $ m_{ref} \approx $ 4, 13, 24 [kg] for the feet, lower and upper legs, respectively. Assuming the robot's mass as $ m_{rob} =$ 20 [kg], we compute the kinetic energy which a collision would absorb (and possibly restore) as $ \Delta E = \mu v_{rel}^2/2 $, where $ \mu = (m_{ref}^{-1}+m_{rob}^{-1})^{-1} $ is the reduced mass and $ v_{rel} $ is the relative velocity of both agents' centers towards each other. As $ \Delta E $ directly quantifies the bodies' potential deformation, it provides a measure for collisions' severity~\cite{7353110}.

\section{Results}

\subsection{Benchmark results}\label{results}
\begin{figure*}
  \centering
  \setlength{\tabcolsep}{0pt}
  \begin{tabular}{cc}
  \includegraphics[trim=0 45 0 0, clip,width= 0.5\linewidth ]{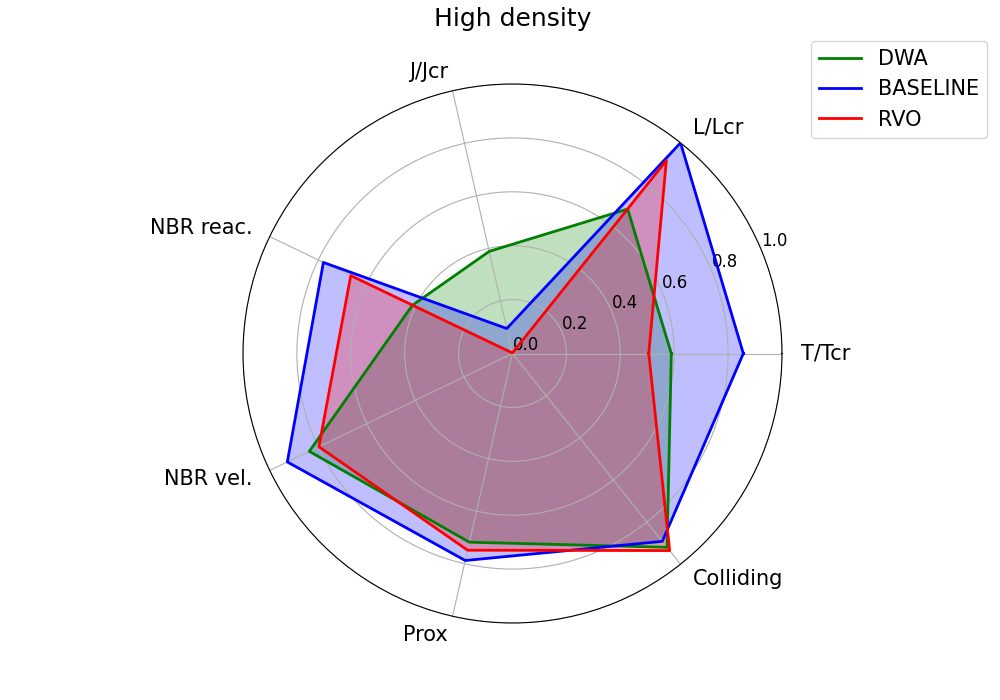} &
  \includegraphics[trim=0 45 0 0, clip,width= 0.5\linewidth ]{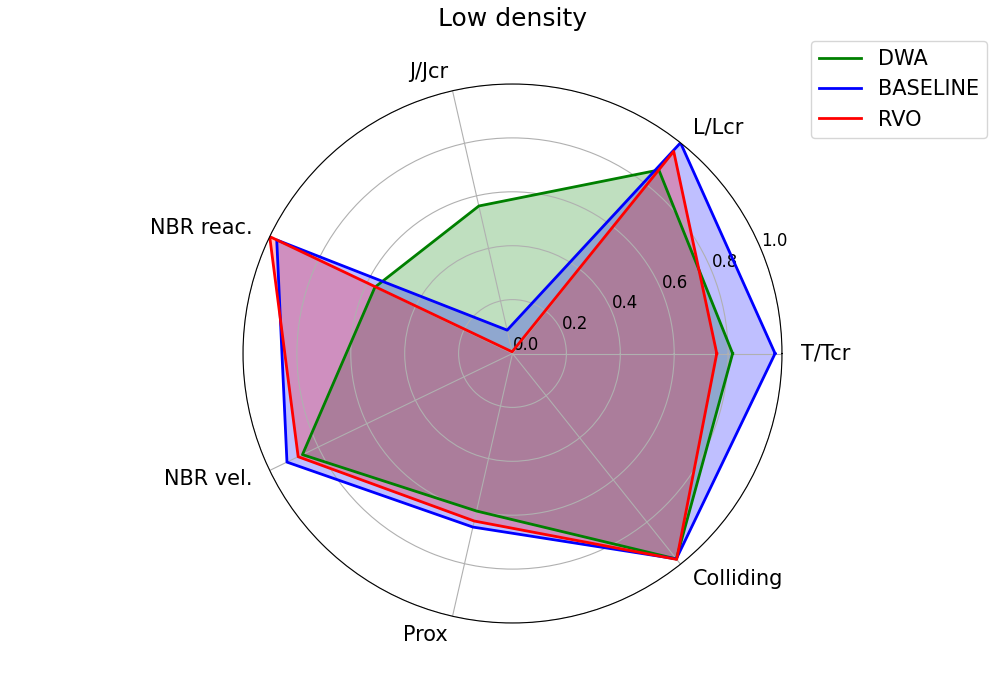} \\
  a) & b)\\
  \end{tabular}
  \caption{\label{fig:radar_charts} Radar charts of the benchmark. (a) We compute the seven metrics described in Section~\ref{metrics} for 20 scenarios: 4 levels of density, 5 crowd flows (see Section~\ref{benchmark}). (b) Similarly, we compute the same radar chart for 5 scenarios with low density only (0.1pm$^{-2}$)}
\end{figure*}

We represent the seven metrics described in Section~\ref{metrics} on two radar charts shown in Fig.~\ref{fig:radar_charts}. Table~\ref{table:radarDeviation1} gives the standard deviation for each metric of the radar charts.

\begin {table}[h]
\centering
\caption{Standard deviation for high and low density radar charts}
\label{table:radarDeviation1}
\resizebox{\columnwidth}{!}
{
\begin{tabular}{|l|c|c|c|c|c|c|c|}
  \hline
  \multirow{2}{*}{Navigation method} & \multicolumn{7}{c|}{Metrics}\\
   & \textit{T/Tcr} & \textit{L/Lcr} & \textit{J/Jcr} & \textit{NBR reac.} & \textit{NBR vel.} & \textit{Prox.} & \textit{Colliding}   \\
  \hline
  BASELINE High density & 0.190 & $0.00$ & $0.238$ & $1.022$ & $0.070$ & $0.132$ & $0.132$\\
  BASELINE Low density & 0.022 & $0.00$ & $0.244$ & $1.402$ & $0.084$ & $0.164$ & $0.027$\\
  \hline
  DWA High density & $0.257$ & $0.225$ & $0.289$ & $0.488$ & $0.089$ & $0.175$ & $0.117$\\
  DWA Low density & $0.171$ & $0.123$ & $0.280$ & $0.0.807$ & $0.096$ & $0.171$ & $0.040$\\
  \hline
  RVO High density & $0.236$ & $0.100$ & $0.015$ & $1.052$ & $0.129$ & $0.141$ & $0.092$\\
  RVO Low density & $0.145$ & $0.067$ & $0.029$ & $1.863$ & $0.125$ & $0.149$ & $0.031$\\
  \hline
\end{tabular}
}
\end {table}


\begin{figure}
  \centering
  \setlength{\tabcolsep}{0pt}
  \includegraphics[width= 0.7\linewidth ]{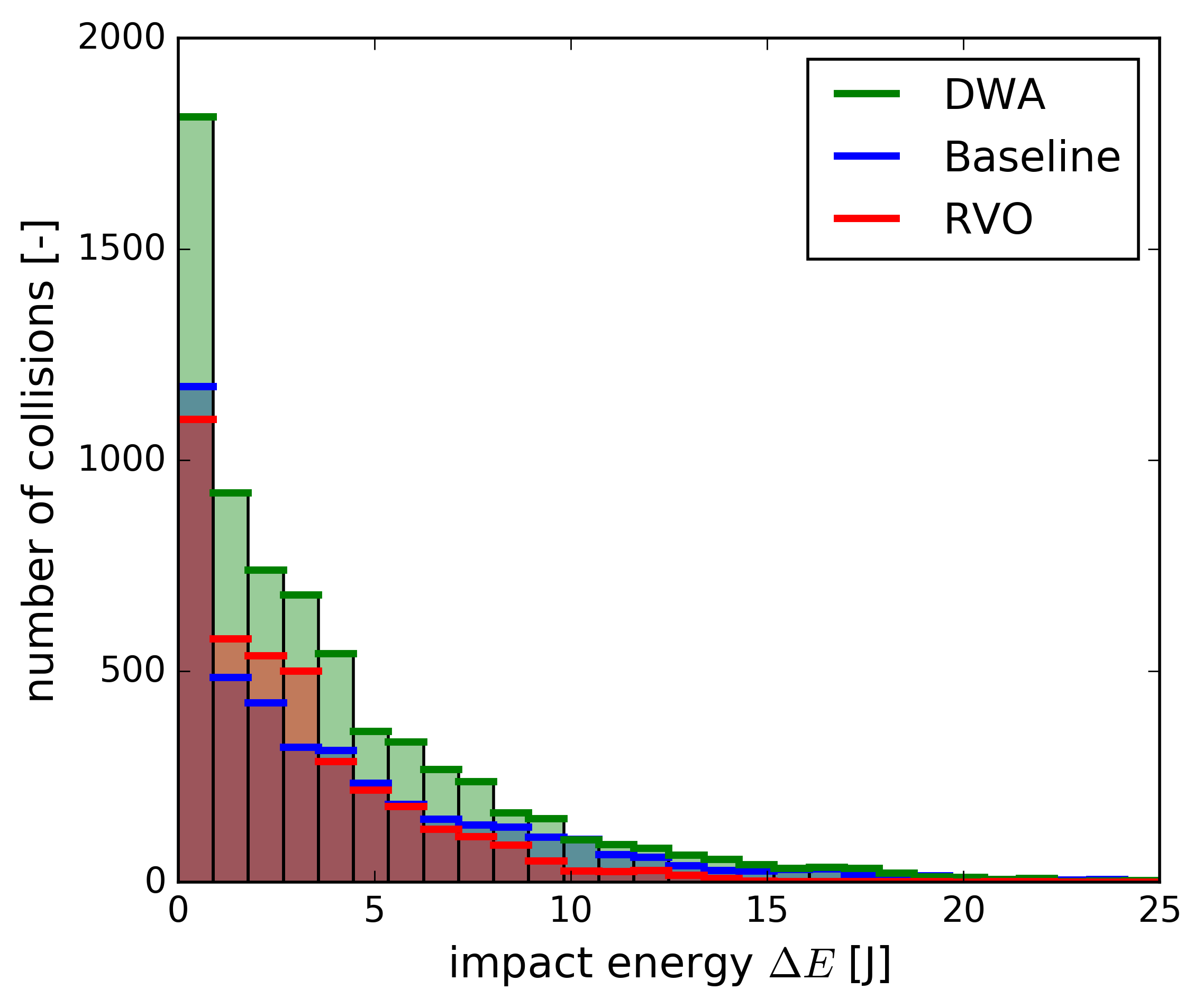}
  \caption{\label{fig:impact_energy} The histograms count the collisions for each robotic controller over all simulations, discriminating them by estimated energy absorption.}
\end{figure}

\subsection{Collision Assessment Results}\label{assessment_results}

Fig.~\ref{fig:impact_energy} shows how many collisions occurred with particular values of energy $ \Delta E $ during all the simulations with a given robotic controller. To normalize by the simulated time, we define a particular robotic controller's collision rate as $ f_c = N_c/T $ and energy rate $ Q = \Sigma\Delta E/T $, where $ N_c $, $ T $ and $ \Sigma\Delta E $ denote respectively the total number of collisions, elapsed time and energy absorption in corresponding simulations. For the baseline, DWA, and RVO, we measure $ f_c =$ 0.624, 0.610, 0.311 [1/s] and $ Q =$ 2.568, 2.285, 0.901 [J/s], respectively.

\section{Discussion}

This section takes each of the results and gives an interpretation in two levels of comparison: path efficiency and crowd safety.

\subsection{Path efficiency}
The path efficiency can be evaluated using the radar chart on Fig.~\ref{fig:radar_charts}.
For the metric $L/Lcr$, an indicator of path length,
naturally, the score for the baseline method is 1, as the method will take the shortest path possible in the corridor (which validates $H1$). RVO is a lot more direct than DWA in both low density and high density crowds, which shows the tendency of DWA to take less direct paths than RVO to reach the goal (validates $H2$ and $H3$). This idea is reinforced by the $J/Jcr$ metric which is an indicator of the variation of linear and rotational speed. Indeed, the baseline and RVO methods have a lot of erratic movements (start and stop), while DWA takes the robot dynamics into account and punishes the difference between the current velocity and desired velocity.
The metric $T/Tcr$, which is an indicator of the time taken to reach the goal, shows that the most direct method, the baseline, is the fastest in both low and high density scenarios (which validates $H1$). The two methods RVO and DWA are much slower in high density scenarios. Indeed, such scenarios lead to situations where the only solution for the robot is to totally stop the movements or even move back, facing the starting point instead of the goal.

\subsection{Crowd safety}
The crowd safety can be appreciated through the Proximity metric, the Colliding metric, the neighbour-related metrics (``NBR reac." and ``NBR vel.") and the histograms of impact energy on Fig.~\ref{fig:impact_energy}.

First, it is interesting to notice that, on the radar chart for low density crowds, the colliding metric has a maximum value for each of the methods which means that in low density (0.1 pm$^{-2}$) the crowd is perfectly capable of avoiding the robot without it using any specific method. In high density scenarios, the baseline has a lower score than the two other methods: the robot does not try to avoid the crowd, thus the risk of collision increases (which validates $H1$). Also, in high density scenarios, RVO performs better than DWA on the colliding metric (which validates $H2$ and $H3$).
The crowd reaction to the robot can be evaluated through the crowd velocity metrics: ``NBR reac.", and ``NBR vel.".
The radar charts show that in high density crowds, the robot's neighbors are more likely to change their orientation instead of their speed. In low density, the robot presence does not impact the crowd much, with the exception of the DWA ``NBR reac." metric which shows that the robot's neighbors change their orientation a lot more than the rest of the crowd. We think that the robot movement generated by RVO and the baseline methods generate more linear trajectories than the DWA method (which validates $H3$). Linear trajectories are easier to predict by the crowd simulators we used in the benchmark (which, at best, linearly extrapolate the trajectory).
The Proximity is naturally higher in high density than in low density crowds, but the robot navigation method does not seem to affect this metric.
The histograms of impact energy in Fig.~\ref{fig:impact_energy} show that the 3 considered techniques generate a similar distribution of impact energies and thus also impact forces when collisions occur. It is also visible that DWA leads to the most collisions at all energy levels. However, the larger total number results from longer simulations, as the rate of collision $ f_c $ (Sec.~\ref{assessment_results}) for DWA is in between both other methods (confirming the colliding metric's ranking).  
Still, the Baseline collides more often than RVO on high energy, while RVO collide more often than the baseline on low energy, which makes it a good compromise between the number of collision and energy (which validates $H1$ and $H3$).
Finally, the rates of collision and energy absorption $ f_c $ and $ Q $ (Sec.~\ref{assessment_results}) are consistently bad for the Baseline, better for DWA, and a lot better for RVO, which validates $H1$, $H2$ and $H3$ and indicates that RVO leads to the lowest number and risk of collisions occurring per unit time.


\section{Conclusion}

We have introduced a new software to simulate the navigation of robots in virtual crowds. We have demonstrated the potential of such simulator to be used as a benchmark tool to compare various navigation techniques. This paper demonstrate limited results, only sufficient to demonstrate the relevance of this solution to be used as a benchmark tool. Nevertheless, we believe that the solution we propose is of interest for the community and opens new perspective in the evaluation of robot crowd navigation capabilities. For future work, we want to first extend the metrics by which we evaluate the robot behaviour. For example, we would like to measure near-collision situations to evaluate more deeply safety aspects. Other elements than just trajectories should be considered. For example, simulation can further explore the question of sensing the crowd such as by comparing navigation results in various sensors configuration. Another direction is to extend the number of scenarios we consider. The strength of simulation is its potential to explore many situation for the robot, and is yet unexploited here. Devising scenarios of greatest interest should however be done at the community scale.


\section*{ACKNOWLEDGMENT}

This project has received funding from the European Union's Horizon 2020 research and innovation programme under grant agreement No 779942, CrowdBot.

\bibliographystyle{IEEEtran}
\bibliography{references}

\end{document}